\documentclass{article}
\pdfoutput=1
\usepackage{pdfpages}
\usepackage{arxiv}
\usepackage{graphicx}
\usepackage[utf8]{inputenc} 
\usepackage[T1]{fontenc}    
\usepackage{url}            
\usepackage{booktabs}       
\usepackage{amsfonts}       
\usepackage{nicefrac}       
\usepackage{microtype}      
\usepackage{algpseudocode}
\usepackage{verbatim}
\usepackage{mathptmx}
\usepackage{mathtools}
\usepackage{amsmath}
\usepackage{xcolor}
\usepackage{float}
\floatstyle{plaintop}
\restylefloat{table}
\usepackage{caption} 
\usepackage{array}
\usepackage{makecell}
\usepackage{caption}

\usepackage[ruled]{algorithm2e}

\usepackage{amsmath,amsfonts,bm}

\def\eqref#1{equation~\ref{#1}}

\def\1{\bm{1}}

\def\vb{{\bm{b}}}
\def\vc{{\bm{c}}}

\def\vs{{\bm{s}}}

\def\vx{{\bm{x}}}

\def\vz{{\bm{z}}}

\def\mA{{\bm{A}}}

\def\mS{{\bm{S}}}

\def\mU{{\bm{U}}}
\def\mV{{\bm{V}}}
\def\mW{{\bm{W}}}

\def\mZ{{\bm{Z}}}

\def\mSigma{{\bm{\Sigma}}}

\DeclareMathAlphabet{\mathsfit}{\encodingdefault}{\sfdefault}{m}{sl}
\SetMathAlphabet{\mathsfit}{bold}{\encodingdefault}{\sfdefault}{bx}{n}

\newcommand{\R}{\mathbb{R}}

\begin{document}

\title{Reduced-Order Modeling of Deep Neural Networks}

\author{
Julia Gusak\textsuperscript{\thanks{Contributed equally.}},
Talgat Daulbaev\textsuperscript{$^*$},
Evgeny Ponomarev,
Andrzej Cichocki,
Ivan Oseledets \\
Skolkovo Institute of Science and Technology \\
\texttt{\{y.gusak, t.daulbaev, evgenii.ponomarev, a.cichocki, i.oseledets\}@skoltech.ru}
}

\date{}





\maketitle

\begin{abstract}
We introduce a new method for speeding up the inference of deep neural networks. 
It is somewhat inspired by the reduced-order modeling techniques for dynamical systems.
The cornerstone of the proposed method is the maximum volume algorithm.
We demonstrate efficiency on neural networks pre-trained on different datasets. 
We show that in many practical cases it is possible to replace convolutional layers with much smaller fully-connected layers with a relatively small drop in accuracy.
\end{abstract}
\section{Introduction}
Recent studies~\cite{chen2018neural,grathwohl2018ffjord,gusak2020towards,daulbaev2020interpolated} have shown the connection between deep neural networks and systems of ordinary differential equations (ODE). 
In these works, the output of the layer during the forward pass was treated as the state of a dynamical system at a given time.
One of the effective methods for accelerating computations in dynamical systems is the construction of reduced models~\cite{quarteroni2014reduced}. 
The classical approach for building such models is the Discrete Empirical Interpolation Method (DEIM; see~\cite{chaturantabut2010nonlinear}).
The idea of DEIM is based on a low-dimensional approximation of the state vector, combined with efficient recalculation of the coefficients in this low-dimensional space through the selection of the submatrix of sufficiently large volume. 

In this work, we use the above connection to build a reduced model of deep neural network for a given pre-trained (fully-connected or convolutional) network. 
We call this model Reduced-Order Network (RON).  
The reduced model is a fully-connected network that has smaller computational complexity than the original neural network. The complexity is defined as the number of floating-point operations (FLOP) required to propagate through the network.
Thus, the inference of RON can be faster.

Following the reduced-order modeling approach, we assume that the outputs of some layers lie in low-dimensional subspaces. 
We will refer to this assumption as the \emph{low-rank assumption}.
Let $\vx$ be the object from the dataset, and $\vz_k = \vz_k(\vx)$ be the vectorized output of the $k$-th layer. 
We assume that there exists a matrix 
$\mV_k \in \mathbb{R}^{D_k \times R_k} \ (D_k \gg R_k)$ 
such that
\begin{equation}
    \vz_k \cong \mV_k \vc_k,
\end{equation}
where $\vc_k = \vc_k(\vx)$ are embeddings. 
The matrix $\mV_k$ is the same for all $\vx$.  

This simple linear representation itself can not help to reduce the complexity of neural networks, because all linear operations in a neural network are followed by non-linear element-wise functions. 
However, we propose how to approximate the next embedding based on the previous one.

As a result, under the low-rank assumption most fully-connected and convolutional neural networks\footnote{We mean convolutional neural networks consisting of convolutions, non-decreasing activation functions, batch normalizations, maximum poolings, and residual connections.}
can be approximated by fully-connected networks with a smaller number of processing units. 
In other words, instead of dealing with huge feature maps, we project the input of the entire network into a low-dimensional space and then operate with low-dimensional representations. We restore the output dimensionality of the model using a linear transformation. 
As a result, the complexity of neural networks can be significantly decreased. 

Even if the low-rank assumption holds only very approximately, we still can use it to initialize a new network and then perform several iterations of fine-tuning. 

Our main contributions are:
\begin{itemize}
\itemsep0em 
\item We propose a new low-rank training-free method for speeding up the inference of pre-trained deep neural networks and show how to efficiently use the rectangular maximum volume algorithm to reduce the dimensionality of layers and estimate the approximation error.
\item We validate and evaluate performance the proposed approach in a series of computational experiments with LeNet pre-trained on MNIST and VGG models pre-trained on CIFAR-10/CIFAR-100/SVHN.
\item We show that our method works well on top of pruning techniques and allows us to speed up the models that have already been accelerated.
\end{itemize}
\section{Related work}
\label{related_work}

Recently, a series of approaches have been proposed to speed up inference in convolutional neural networks~(CNNs)~\cite{Cheng2018}. In this section, we overview the main ideas of different method families and highlight the differences between them and our approach. 

Many different methods deal with a pre-trained network, which we call the \textit{teacher} network, and an accelerated network, which we call the \textit{student} network. 
This terminology comes from \textbf{knowledge distillation}~\cite{Bucilua2006,Hinton2015,Romero2014,Zagoruyko2016_attention} methods, where the softmax outputs of the teacher network are used as a target vector for the student network.

In section~\ref{sec::compare}, we compare our aproach with different \textbf{channel pruning} methods.
Such methods aim to prune redundant channels in the weight tensors and, hence, accelerate and compress the whole model.
Pruned channels are selected due to special information criteria. 
For example, it can be a sum of absolute values of weights~\cite{li2016pruning} or an average percentage of zeros~\cite{hu2016network}.

There are two major approaches to channel pruning. 
The first approach is to deal with a single network and train it from scratch, adding extra regularization, which forces the channel-level sparsity of weights. 
Later on, some channels are considered to be redundant and have to be removed~\cite{Liu2017learning,wen2016learning}.
It is usually an iterative procedure, which is computationally expensive, especially for very deep neural networks.
The second approach involves both teacher and student networks. 
The student is trained to minimize the reconstruction error between feature maps of two models~\cite{he2017channelpruning,hu2016network,ThiNet}. 

In \cite{he2017channelpruning}, channel selection is made using LASSO regression, and the reconstruction is performed via least squares. In~\cite{ThiNet}, the pruning strategy for a layer depends on the statistics of the next layer. 
In~\cite{Liu2017learning}, it is proposed to multiply each channel on a unique learnable scalar parameter; then, the whole network is trained with a sparsity regularization on these scalar parameters.
In~\cite{zhong2018prune}, neural architecture search techniques are combined with channel pruning. Namely, the pruning strategy is generated by the LSTM network, which is trained in a reinforcement learning way. 

In~\cite{zhuang2018discrimination}, Discrimination-aware Channel Pruning (DCP) algorithm is introduced. 
It is a multi-stage pruning method applied to the pre-trained network. 
At each stage of DCP, a network from the previous stage is trained with an additional classifier and discriminative loss. The least informative channels either pruned at a fixed rate or selected using a greedy algorithm. 
We refer to these approaches as to DCP and DCP-Adapt, accordingly.

Another related family of acceleration methods is \textbf{low-rank methods}, which uses matrix or tensor decomposition to estimate the informative parameters of deep neural networks. 
In most cases, a much lower total computational cost can be achieved by replacing a convolutional layer with a sequence of several smaller convolutional layers~\cite{Denton2014,Jaderberg2014,Lebedev2014,Zhang2015,gusak2019automated}.
Opposed to our approach, most low-rank methods are applied not to feature maps~\cite{cui2019active} but to weight tensors. 

Finally, \textbf{quantization}~\cite{Courbariaux2014,Gupta2015} methods worth mentioning. 
Such methods can significantly accelerate networks, but they usually require special hardware to reach a theoretical speed-up in practice.

\section{Background}
In this section, we give a brief description of the rectangular volume algorithm (Subsection~\ref{sec:maxvol}) and explain how to compute low-dimensional subspaces of embeddings (Subsection~\ref{sec:V}). This information is required to clearly understand what follows.

\subsection{Maximum Volume Algorithm and Sketching}
\label{sec:maxvol}
The rectangular maximum volume algorithm~\cite{mikhalev2018rectangular} is a greedy algorithm that searches for a maximum volume submatrix of a given matrix. The volume of a matrix $\mA$ is defined as
\begin{equation}
    \mathrm{vol} \left( A \right)
        =
    \det( \mA^\top \mA ).
\end{equation}
This algorithm has several practical applications~\cite{fonarev2016efficient,mikhalev2018rectangular}. 
In this paper, we use it to reduce the dimensionality of overdetermined systems as follows.

Assume, $\mA \in \R^{D \times R}$ is a tall-and-thin matrix ($D \gg R$); and we have to solve a linear system
\begin{equation}
\label{eq:linsys}
    \mA \vx = \vb
\end{equation}
with a fixed matrix $\mA$ for an arbitrary right-hand side $\vb \in \mathbb{R}^{D}$. 
The solution is typically given by
\begin{equation}
    \vx = \mA^\dagger \vb,
\end{equation}
where $\mA^\dagger = (\mA^\top \mA)^{-1} \mA^\top$ is the Moore-Penrose pseudoinversion of $\mA$. 
The issue is that a matrix-by-vector product with $R \times D$ matrix $\mA^\dagger$ costs too much. 
Moreover, for the ill-conditioned matrix, the solution is not very stable. 

Instead of using all $D$ equations, we can select the most ``representative'' of them.
For this purpose, we apply the rectangular maximum volume algorithm\footnote{\url{https://bitbucket.org/muxas/maxvolpy}} to the matrix $\mA$. 
It returns a set $P$ row indices ($R \le P \ll D$), which corresponds to equations used for further calculations. In this work, we choose $P$ on the segment $[R, 2R]$.

A submatrix consisting of $P$ given rows can be viewed as $\mS \mA$, where $\mS \in \lbrace 0, 1 \rbrace^{P \times D}$.
We call $\mS$ a \textit{sketching matrix}. 
For convenience in notations, we assume that the rectangular maximum volume algorithm outputs a sketching matrix. 
Thus, the system~(\ref{eq:linsys}) can be solved as follows
\begin{equation}
    \vx = (\mS \mA)^\dagger \left( \mS \vb \right). 
\end{equation}
Selecting rows of $\vb$ is a cheap operation, so the complexity of computing $\mS \vb$ is $O(P)$.
If $(\mS \mA)^\dagger$ is precomputed, for any right-hand side we only have to carry out matrix-by-vector multiplication with a matrix of size $R \times P$.

\subsection{Computation of Low-Dimensional Embeddings}
\label{sec:V}

Let $\mZ \in \R^{N \times D}$ be the output matrix of a given layer; each row of $\mZ$ corresponds to a training sample propagated through the part of the network ending with this layer. 

The truncated rank-$R$ SVD of $\mZ^\top \in \R^{D \times N}$ is given by 
\begin{equation}
    \mZ^\top 
        \cong
    \underbrace{\mV}_{D \times R}
    \ 
    \underbrace{\mSigma \mU^\top}_{R \times N}.
\end{equation}

Here the matrix $\mV$ corresponds to the linear transformation, which maps to the low-dimensional embedding subspace.
To compute the matrix $\mV$, we use the matrix sketching algorithm based on hashing~\cite{woodruff2014sketching,tsitsulin2020frede}.
For our applications, it is faster than randomized SVD. 
\section{Method}

Our goal is to build an approximation of a given deep neural network (\textit{teacher}) by another network (\textit{student}) with much faster inference. 

Most conceptual details of our approach are explained on a toy example of a multilayer perceptron (Subsection~\ref{sec:mlp}). 
Later on, we describe how to apply the proposed ideas to feed-forward convolutional neural networks (Subsection~\ref{sec:cnn}) and residual networks (Subsection~\ref{sec:resnet}).

\subsection{A Toy Example: MLP}

In this subsection, we consider a simple fully-connected feed-forward neural network, or multilayer perceptron (MLP).

Hereinafter let $\psi_k$ ($k = 1, \ldots, K$) be non-decreasing element-wise activation functions, e.g., ReLU, ELU or Leaky ReLU. 
Note that our method allows us to accelerate a part of the initial network, but for simplicity, we assume that the whole teacher network is used. 
Besides, without loss of generality, we suppose that all biases are equal to zero.

Let $\vz_0$ be an input sample.
Being passed through $K$ layers of the teacher network, it undergoes the following transformations
\begin{equation}
    \vz_1 = \psi_1 (\mW_1 \vz_0), \
    \vz_2 = \psi_2 (\mW_2 \vz_1), \
    \ldots, \ 
    \vz_K = \mW_K \vz_{K - 1},
    \label{eq:teacher}
\end{equation}
where $\mW_k \in \R^{D_{k} \times D_{k - 1}}$ is a weight matrix of the $k$-th layer.

Let $\vc_1, \ldots, \vc_K$ be the embeddings of $\vz_1, \ldots, \vz_K$. 
We have already known how to compute the linear transformation $\mV_k \in \R^{D_k \times R_K}$, which maps $\vz_k$ to $\vc_k$.
Here the dimensionality of the $k$-th embedding $R_K$ is much smaller than the number of features $D_k$.

The low-rank assumption for the first layer gives
\begin{equation}
    \vz_1 
\cong 
\boxed{
    \mV_1 \vc_1
\cong
    \psi_1 (\mW_1 \vz_0)
}
\end{equation}

The boxed expression is a tall-and-skinny linear system with the matrix $\mV_1 \in \R^{D_1 \times R_1}$, the right-hand side vector $\psi_1 (\mW_1 \vz_0)$ and the vector of unknowns $\vc_1$. If $\mS_1 \in \R^{P_1 \times D_1}$ is a sketching matrix (Section~\ref{sec:maxvol}) for the matrix $\mV_1$, we can compute the embedding as follows
\begin{equation}
    \vc_1
\cong 
    \left(\mS_1 \mV_1 \right)^\dagger 
    \mS_1 \psi_1 \left( 
        \mW_1 \vz_0
    \right)
=
    \underbrace{\left(\mS_1 \mV_1 \right)^\dagger}_{R_1 \times P_1} 
    \psi_1 ( 
        \underbrace{\mS_1 \mW_1}_{P_1 \times D_1} 
        \vz_0
    ).
\end{equation}
Here we switch point-wise linearity $\psi$ and sampling because they commute pairwise.

The same technique can be applied for computing the second embedding $\vc_2$ using $\vc_1$. We write the low-rank assumption
\begin{equation}
    \vz_2
\cong 
    \psi_2 \left(\mW_2 \vz_1 \right)
\cong 
    \psi_2 \left(\mW_2 \mV_1 \vc_1 \right)
\cong 
    \mV_2 \vc_2,
\end{equation}
get the linear system
\begin{equation}
\boxed{
    \mV_2 \vc_2
\cong
    \psi_2 \left(\mW_2 \mV_1 \vc_1 \right)}
\end{equation}
and apply the rectangular maximum volume algorithm. 
If $\mS_2 \in \R^{P_2 \times D_2}$ is a sketching matrix, $\vc_2$ can be estimated as 
\begin{equation}
    \vc_2 
\cong 
    \underbrace{\left(\mS_2 \mV_2 \right)^\dagger}_{R_2 \times P_2} 
    \psi_2 ( 
        \underbrace{\mS_2 \mW_2 \mV_1}_{P_2 \times R_1} 
        \vc_1 
    ).
\end{equation}

The process can be continued for other layers.
The output of the student network is computed as $\mV_K \vc_K$:
\begin{align}
\begin{split}
\vc_1 &\cong \underbrace{\left(\mS_1 \mV_1 \right)^\dagger}_{R_1 \times P_1} 
    \psi_1 (
        \underbrace{\mS_1 \mW_1}_{P_1 \times D_1} 
        \vz_0
    ) 
\\
&\quad \quad \quad \quad \ldots \\
\vc_k &\cong \underbrace{\left(\mS_k \mV_k \right)^\dagger}_{R_K \times P_k} 
    \psi_2 ( 
        \underbrace{\mS_k \mW_k \mV_{k - 1}}_{P_k \times R_{k - 1}} 
        \vc_{k - 1}
), \quad k = 1, \ldots, K
\\
\vz_K &\cong \mV_K \vc_K
\end{split}
\label{eq:c_system}
\end{align}
Suppose $\vs_k$ is the output of $\psi_k$.
We can rewrite~(\ref{eq:c_system}) in a better way
\begin{align}
\begin{split}
\vs_1 &\cong  
    \psi_1 (
        \underbrace{\mS_1 \mW_1}_{P_1 \times D_1} 
        \vz_0
    ), 
\\
\vs_2 &\cong  
    \psi_2 ( 
        \underbrace{\mS_2 \mW_2 \mV_1 
        \left(\mS_1 \mV_1 \right)^\dagger}_{P_2 \times P_1} \vs_1 
),  
\\
&\quad \quad \quad \quad \ldots \\
\vs_K &\cong
    \psi_K ( 
        \underbrace{\mS_K \mW_K \mV_{K - 1} 
        \left(\mS_{K - 1} \mV_{K - 1} \right)^\dagger}_{P_K \times P_{K - 1}} \vs_{K - 1}
),
\\
\vz_K &\cong \underbrace{\mV_K \left(\mS_K \mV_K \right)^\dagger}_{D_k \times R_K} \vs_K.
\end{split}
\label{eq:s_system}
\end{align}
As a result, instead of $K$-layer network with $D_k \times D_{k + 1}$ layers~(\ref{eq:teacher}) we obtain a more compact $K + 1$-layer network~(\ref{eq:s_system}).  

The proposed approach is summarized in Algoritm~\ref{algo:mlp}.

Then, we propose to add batch normalizations into the accelerated model and perform several epochs of fine-tuning. 

\begin{algorithm}
\DontPrintSemicolon 
\KwIn{teacher's weights $\{\mW_1, \mW_2, \ldots, \mW_K\}$ and element-wise activation functions $\{\psi_1, \psi_2, \ldots, \psi_K\}$; subset of the training set $\mZ$~--- a number of samples $\times$ number of input features matrix; $\{R_1, R_2, \ldots, R_K\}$~--- sizes of the embeddings;
}
\KwOut{student's weights $\{ \widetilde{\mW}_0, \widetilde{\mW}_1, \widetilde{\mW}_2, \ldots, \widetilde{\mW}_K\}$;}
\Comment{For simplicity, we use all $\lbrace \mV_k \rbrace_{k = 1}^{K}$, but in fact we have to keep only two of them to compute a single weight of student.}\;
\For{$k \gets 1$ \textbf{to} $K$} {
$\mZ \gets \mZ \text{ propagated through the $k$-th layer}$\;
$\mU, \mSigma, \mV_k \gets \mathrm{truncated\_svd}(\mZ^\top, R_K)$\;
\Comment{In practice, we don't store the whole $\mZ$, but use streaming randomized SVD algorithms}.\;
$\mS_k \gets \mathrm{rect\_max\_vol}(\mV_k)$\;
} 
$\widetilde{\mW}_0 \gets \mS_1 \mW_1$\;
\For{$k \gets 1$ \textbf{to} $K - 1$} {
$\widetilde{\mW}_k \gets \mS_k \mW_k \mV_{k-1} 
        \left(\mS_{k-1} \mV_{k-1} \right)^\dagger$\;
}
$\widetilde{\mW}_K \gets \mV_K \left(\mS_K \mV_K \right)^\dagger$\;
\Return{$\{ \widetilde{\mW}_0, \widetilde{\mW}_1, \widetilde{\mW}_2, \ldots, \widetilde{\mW}_K\}$}\;
\caption{Initialization of the student network}
\label{algo:mlp}
\end{algorithm}

\label{sec:mlp}

\subsection{Convolutional Neural Networks}
\label{sec:cnn}

\textbf{Convolution} is a linear transformation.
We treat it as a matrix-by-vector product, and we convert convolutions to fully-connected layers. Two crucial remarks for this approach should be discussed.

Firstly, we vectorize all outputs. Do we lose the geometrical structure of the feature map? Only partially, because it is integrated into the initial weight matrices.

Secondly, the size of a single convolutional matrix is larger than the size of its kernel. However, these sizes can be compatible after compression if the number of channels is not big. 
So, as a result, a student model can be not only faster but even smaller than the teacher. 

\textbf{Batch normalization} can be merged with the dense layer for inference. 
Thus, in the student model, we get rid of batch normalization layers but preserve the normalization property.

\textbf{Maximum pooling} is a local operation, which typically maps $2 \times 2$ region into a single value~--- the maximum value in the given region.
We manage this layer by taking $4$ times more indices and by applying maximum pooling after sampling.

\subsection{Residual Networks}
\label{sec:resnet}
Residual networks~\cite{he2016deep,zagoruyko2016wide,huang2017densely} are popular models used in many modern applications. In contrast to standard feed-forward CNNs, they are not sequential.
Such models have several parallel branches, the outputs of which are summed up and propagated through the activation function.

We approximate the output of each branch and the result as follows 
\begin{equation}
\mV \vc
    \cong 
\psi \left( 
    \mV_1 \vc_1 + \ldots + \mV_k \vc_k
\right).
\end{equation}

The above expression is an overdetermined linear system. 
If $\mS$ is a sampling matrix for matrix $\mV$, the embedding $\vc$ is computed as
\begin{equation}
\vc 
    \cong 
\left(\mS \mV \right)^\dagger 
\psi \left(
    \mS \mV_1 \vc_1 + \ldots + \mS \mV_k \vc_k
\right).
\end{equation}

The rest steps of residual network acceleration are the same as for the standard multilayer perceptron (Section~\ref{sec:mlp}).

\subsection{Approximation error}

Suppose 
\(
    \mathbf{\varepsilon}_k = \mV_k \vc_k - \vz_k
\)
is an error of the low-rank approximation, thus
\begin{equation}
    \mS_k \mV_k \vc_k
    =
    \left( \mS_k \mV_k \right)^\dagger \mS_k \vz_k + \left( \mS_k \mV_k \right)^\dagger \mS_k \mathbf{\varepsilon}_k.
\end{equation}
and error of our algorithm equals to $e_k = \|\left( \mS_k \mV_k \right)^\dagger \mS_k \mathbf{\varepsilon}_k\|_2$.
Since $\|\mV_k^\top\|_2 = \|\mS_k\|_2 = 1$, 
\begin{equation}
\begin{aligned}
    \| \left( \mS_k \mV_k \right)^\dagger \mS_k \|_2 &= \|\mV_k^\top \mV_k \left( \mS_k \mV_k \right)^\dagger \mS_k\|_2 \\
    &\le 
    \| \mV_k \left( \mS_k \mV_k \right)^\dagger \|_2.
\end{aligned}
\end{equation}

Due to the Lemma 4.3 and Remark 4.4 from the rectangular maximum volume paper~\cite{mikhalev2018rectangular}\footnote{In this paper, the given matrix is defined by $C$.} 
\begin{equation}
\| \mV_k \left( \mS_k \mV_k \right)^\dagger \|_2
    \le 
\sqrt{1 + \dfrac{\left(D_k - P_K \right) r_k}{P_K + 1 - R_K}}.
\end{equation}
Hence,
\begin{equation}
    e_k \le \sqrt{1 + \dfrac{\left(D_k - P_K \right) R_K}{P_K + 1 - R_K}}\|\varepsilon_k\|_2.
\end{equation}
For example, if $P_K = 1.5 R_K$, approximation error $e_k$ is $O(\sqrt{D_k}||\varepsilon_k||_2)$ for $R_K = o(D_k)$.
\section{Experiments}
Firstly, we provide empirical evidence that supports our low-rank assumption about the outputs of some layers.  Secondly, we show the performance of RON for both fully-connected and convolutional neural network architectures. We compare models accelerated by RON with the baselines from DCP~\protect\cite{zhuang2018discrimination} and \protect\cite{zhong2018prune}.

{\bf Datasets}. 
We empirically evaluate the performance of RON on four datasets, including MNIST, CIFAR-10, CIFAR-100, and SVHN. 
MNIST is a collection of handwritten digits that contains images of size $28\times 28$  with a training set of 60000 examples, and a test set of 10000 examples. 
CIFAR-10 consists of $32\times 32$ color images belonging to 10 classes with 50000 training and 10000 testing samples. 
CIFAR-100 is similar to CIFAR-10, but it contains 100 classes with 500 training images and 100 testing images per class. 
SVHN is a real-world image dataset with house numbers that contains 73257 training and 26032 testing images of size $32\times 32$. 

\subsection{Singular values}
\begin{figure}[!ht]
\begin{minipage}[b]{0.5\linewidth}
\center{\includegraphics[width=1.\linewidth]{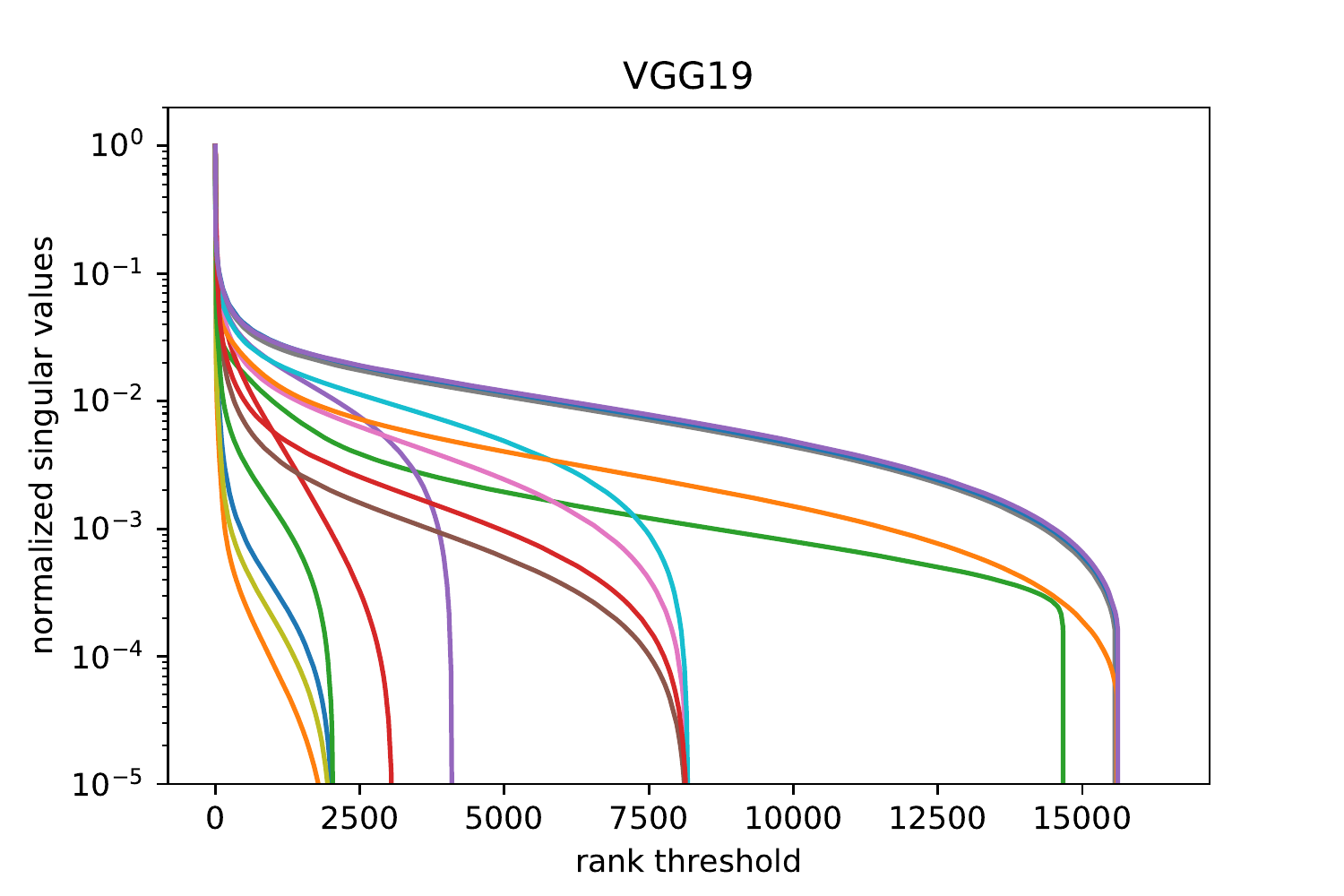}} 
\end{minipage}[b] 
\hfill
\begin{minipage}[b]{0.5\linewidth}
\center{\includegraphics[width=1.\linewidth]{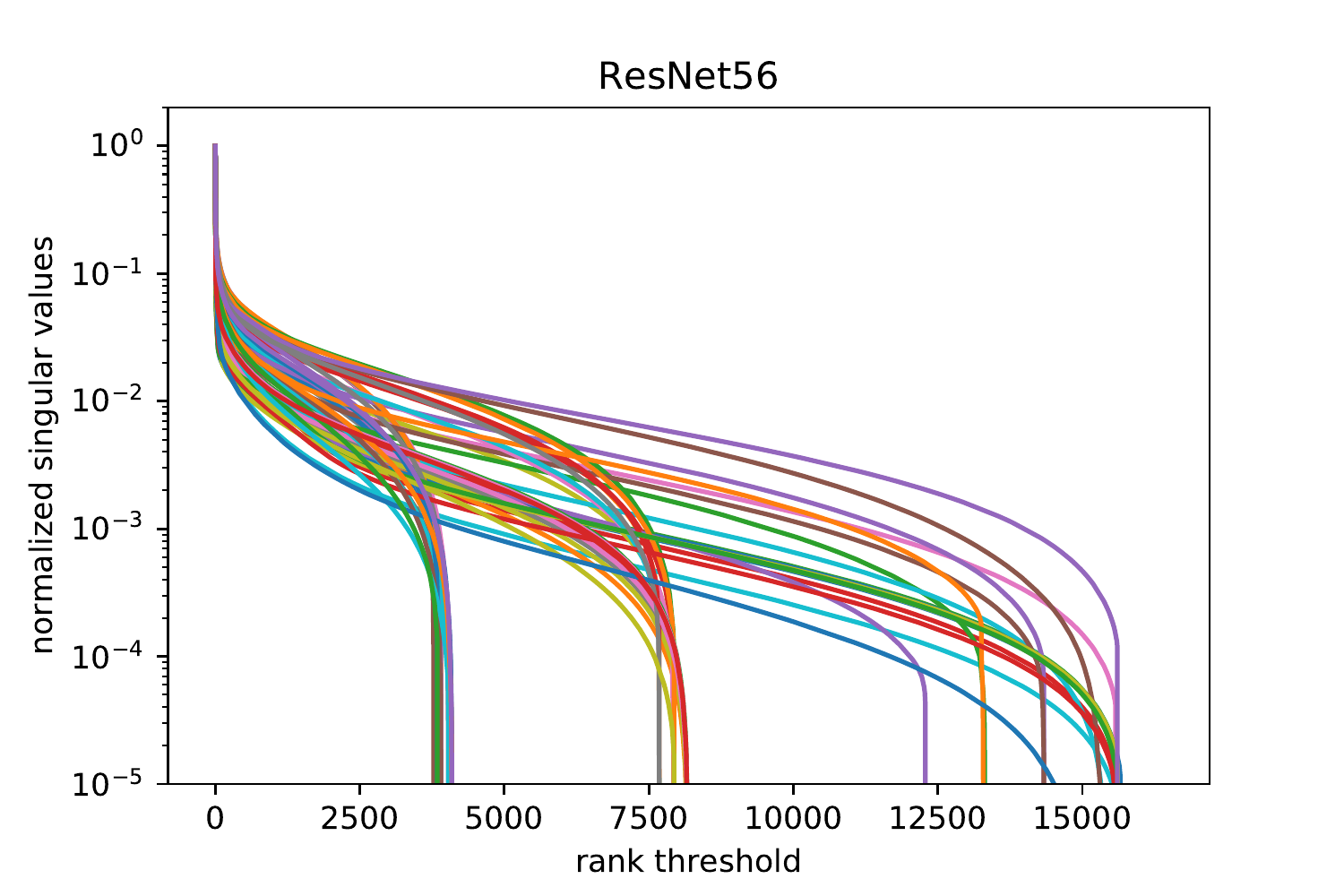}}
\end{minipage}[b]
\caption{We plot singular values of all layers for CIFAR-10. Each singular value is divided by the largest one for this layer. One can see that most singular values are relatively small.}
\label{fig:singularvalues}
\end{figure}

Our method relies on the assumption, which states that the outputs of some layers can be mapped to a low-dimensional space. We perform this mapping using the maximum volume based approximation of the basis obtained through SVD. Figure~\ref{fig:singularvalues} supports the feasibility of our assumption. Each subfigure corresponds to a specific architecture and depicts the singular values of blocks output matrices. It can be seen that the singular values decrease very fast for some (deeper) blocks, which means that their outputs can be approximated by low-dimensional embeddings.

We use two strategies for rank selection: a non-parametric Variational Bayesian Matrix Factorization (VBMF,~\protect\cite{nakajima2013global}) and a simple constant factor rank reduction.

Singular values are computed for matrices containing the whole training data. We use matrix sketching algorithm based on hashing and do not have to store the entire matrix in memory~\protect\cite{woodruff2014sketching,tsitsulin2020frede}.

\subsection{Fully-connected networks}

\begin{figure}[!ht]
\begin{minipage}[b]{0.5\linewidth}
\center{\includegraphics[width=1.\linewidth]{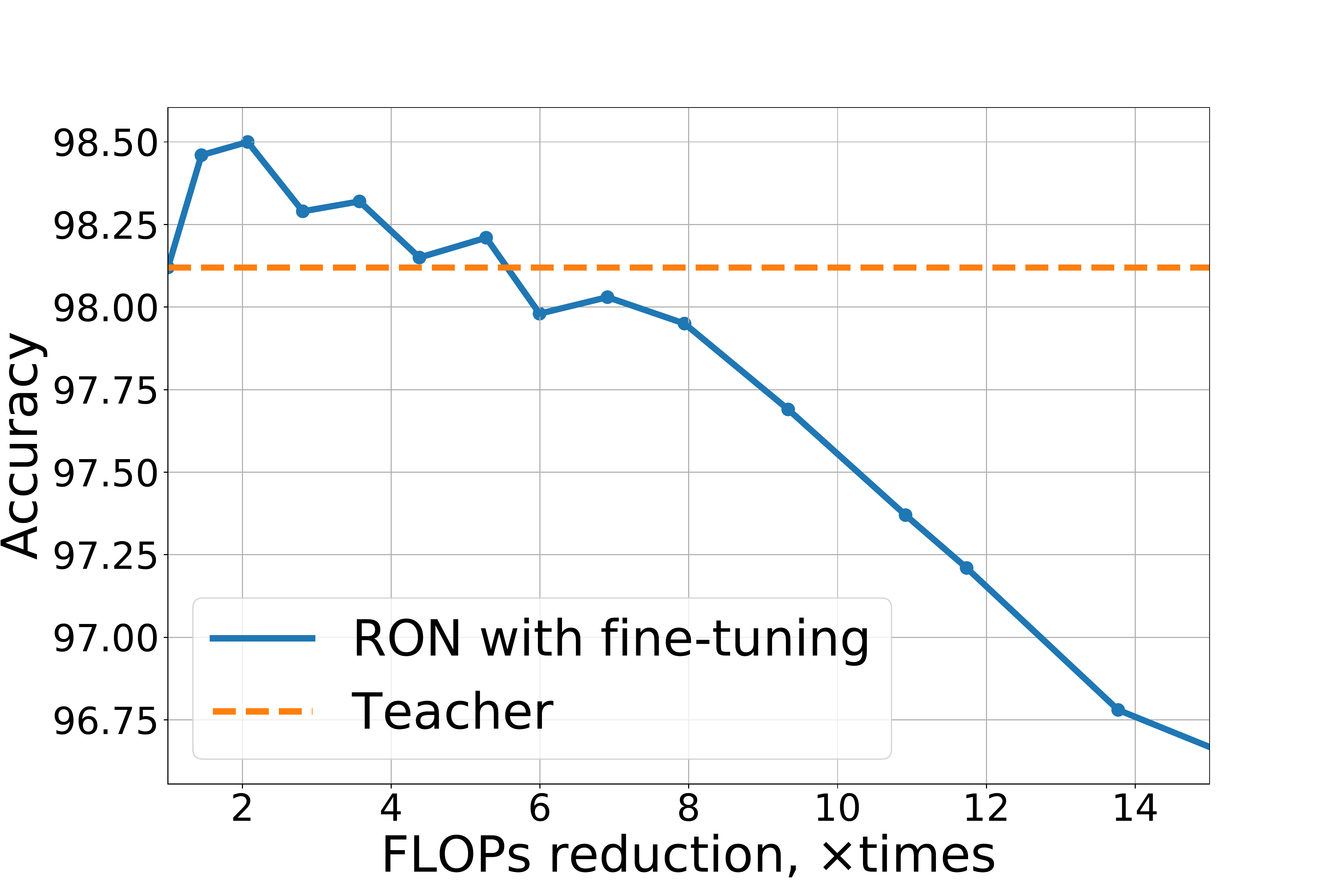}} 
(a) LeNet-300-100-10
\end{minipage}[b] 
\hfill
\begin{minipage}[b]{0.5\linewidth}
\center{\includegraphics[width=1.\linewidth]{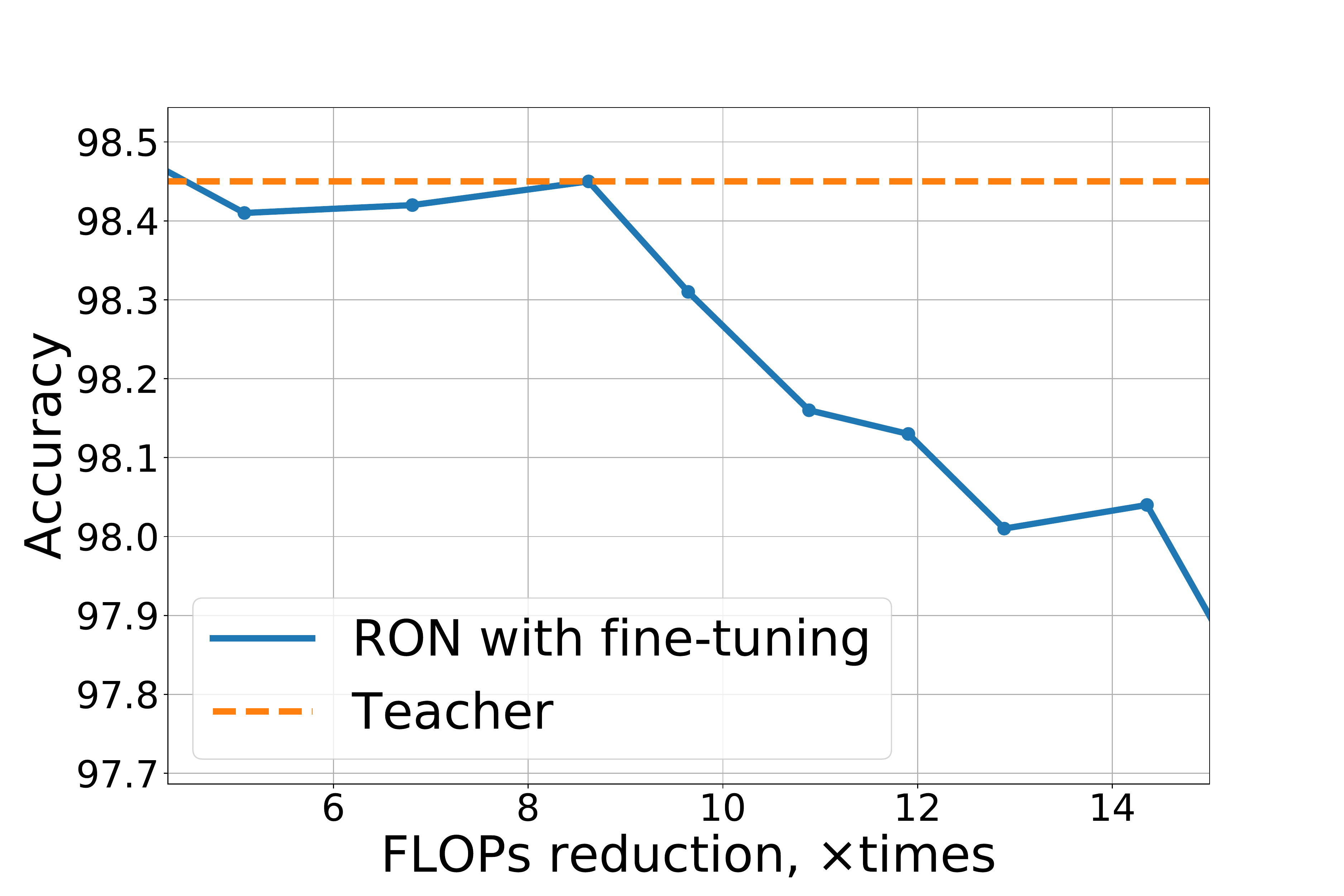}}
(b) LeNet-500-300-10
\end{minipage}[b]
\caption{RON for different LeNet models.}
\label{fig:lenet}
\end{figure}

To illustrate our method, we first choose LeNet-300-100 architecture for MNIST, which is a fully-connected networks with three layers: $784 \times 300$, $300 \times 100$, and $100 \times 10$ with ReLU activations.
We perform 15 iterations of the following procedure.
First, we train the model with the learning rate \texttt{1e-3} for 25 epochs, then we train the model with the learning rate \texttt{5e-4} for the same number of epochs.
After that, we apply our acceleration procedure with rank reduction rates equal to 0.7 and 0.75, respectively.
Figure~\ref{fig:lenet}a shows the FLOP reduction rate together with the test accuracy.

In~\protect\cite{wen2016learning,Liu2017learning,zhong2018prune}, LeNet-500-300 model is accelerated 6.06, 6.41, and 7.85 times, respectively, approximately without accuracy drop.
Our method is able to achieve more then $8\times$ acceleration (see Figure~\ref{fig:lenet}b).

\subsection{Convolutional networks}

We apply our method to VGG-like~\protect\cite{simonyan2014very} architectures on CIFAR-10, CIFAR-100, and SVHN classification tasks. 
In our experiments, we use RON once to obtain an accelerated neural network (student), and then we fine-tune the network if needed. During student initialization (Algorithm~\ref{algo:mlp}), embedding sizes for CIFAR-10 are chosen using VBMF, while for CIFAR-100 and SVHN feature sizes are reduced by a predefined rate. 
Pre-trained teacher models for CIFAR-10 are available online\footnote{\url{https://github.com/SCUT-AILab/DCP/wiki/Model-Zoo}}. 
They include VGG-19 and VGG-19 pruned with DCP~\protect\cite{zhuang2018discrimination} approach at 0.3
For the experiments with CIFAR-100 and SVHN we used pre-trained VGG-19\footnote{\url{https://github.com/bearpaw/pytorch-classification}} and VGG-7\footnote{\url{https://github.com/aaron-xichen/pytorch-playground}} networks, correspondingly.
When applying RON to VGG-like architecture, we accelerate several last convolutional layers of the network. For instance, the model RON (8 to 16) corresponds to the model with nine accelerated convolutional layers.

\paragraph{\textbf{RON without fine-tuning.}} In this setting, we initialize a student network using Algorithm~\ref{algo:mlp} and measure its acceleration and performance.  For VGG-19 on CIFAR-10, RON can achieve $1.53\times$~FLOP reduction with $0.09\%$ accuracy increase without any additional fine-tuning (Table~1). 
We refer to~\protect\cite{zhong2018prune} to provide evidence that RON (without fine-tuning) significantly outperforms one stage of channel pruning~\protect\cite{Liu2017learning,zhong2018prune}(w/o fine-tuning) when applied to the pre-trained VGG-19.
The models with convolutional layers pruned at $0.1\%$ pruning rate (i.e., FLOP reduction is around $1.23\times$) have more than $20\%$ accuracy drop (Figure~5 in~\protect\cite{zhong2018prune}). 

\paragraph{\textbf{RON with fine-tuning.}}
Fine-tuning the model accelerated with  RON, we can achieve $2.3\times$ FLOP reduction with $0.28\%$ accuracy increase (Table~\ref{tab:my_label}) for VGG-19 on CIFAR-10. 
After the acceleration procedure, we perform 250 epochs of fine-tuning by SGD with momentum $0.9$, weight decay \texttt{1e-4}, and batch size 256.
The initial learning rate is equal to \texttt{1e-2}, and it is halved after ten training epochs without validation quality improvement. 
We use dropout during the fine-tuning.

VGG networks for CIFAR-100 (Table~\ref{tab:cifar100}) and SVHN (Table~\ref{tab:svhn}) are less redundant, therefore, acceleration without accuracy drop is smaller than for CIFAR-10 dataset.

Note that compression can be performed iteratively by alternating RON and fine-tuning steps.  
The iterative approach takes much time, but it was shown for both pruning ~\protect\cite{Liu2017learning,zhuang2018discrimination,gao2018dynamic,zhong2018prune} and low-rank~\protect\cite{gusak2019automated} methods that it helps to reduce the accuracy degradation for high compression ratios. 

\begin{table}[htb!]
    \centering
    \begin{tabular}{c|cccc}
        \thead{Model \\ ~} & \thead{Modified \\ layers} &
        \thead{Acc@1 without \\ fine-tuning} & \thead{Acc@1 with \\ fine-tuning} & \thead{FLOP \\ reduction} \\
        \hline
        Teacher & --- & --- &  93.70 &  1.00$\times$ \\
        RON & 10 to 16 & {\bf 93.79} & {\bf 94.10} & {\bf 1.53$\times$}\\
        RON & 9 to 16 & 93.46& {\bf 94.15} & {\bf 1.68$\times$}\\
        RON & 8 to 16 & 90.58 & {\bf 94.24} & {\bf 1.93$\times$}\\
        RON & 7 to 16 & 85.79 & {
        \bf 93.98} & {\bf 2.30}$\times$\\
        RON & 6 to 16 & 72.53 & 93.12 & 3.01$\times$\\
        RON & 5 to 16 & 58.12 & 91.88 & 3.66$\times$\\
        \hline
        DCP~\protect\cite{zhuang2018discrimination} & --- & ---  &  93.96 &  2.00$\times$ \\
        DCP + RON & 10 to 16 & {\bf 93.98  } & {\bf 94.24} & {\bf 3.06$\times$}\\
        DCP + RON & 9 to 16 & {\bf 93.90 } & {\bf 94.27} & {\bf 3.37$\times$}\\
        DCP + RON & 8 to 16 & 91.82 & {\bf 94.01} & {\bf 3.78$\times$}\\
        DCP + RON & 7 to 16 & 88.88 &  {\bf 93.97} & {\bf 4.48}$\times$ \\
        DCP + RON & 6 to 16 & 81.30 & 93.26 & 5.56$\times$\\
        DCP + RON & 5 to 16& 64.12 & 91.5 & 7.21$\times$
        \end{tabular}
    \caption{Accuracy and FLOP trade-off for the models accelerated with RON on CIFAR-10 dataset. DCP is a channel pruning method from~\protect\cite{zhuang2018discrimination}.}
    \label{tab:my_label}
\end{table}

\begin{figure}[!ht]
    \centering
    \includegraphics[width=0.7\linewidth]{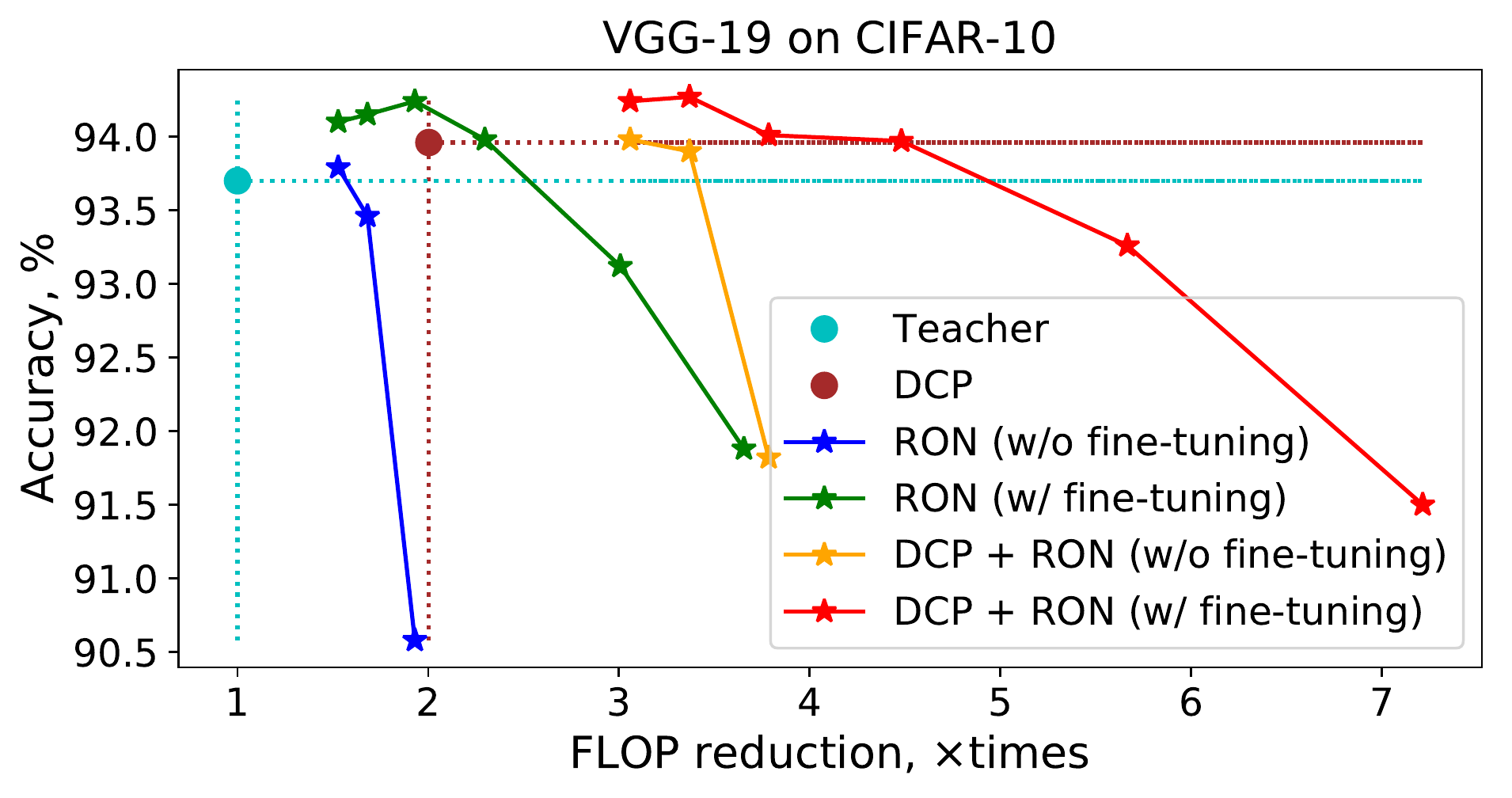}
    \caption{
    Accuracy and FLOP reduction for RON accelerated models on CIFAR-10. 
    }
    \label{fig:cifar10}
\end{figure}

\paragraph{\textbf{RON on top of the pruned network.}}
The motivation to use RON on top of pruned models is the following. Channel pruning methods tend to leave the most informative channels (e.g., in DCP~\protect\cite{zhuang2018discrimination} they look for more discriminative channels) and eliminate the rest. However, a convolutional layer consisting of informative channels can still have a low-rank structure and, therefore, can be further accelerated using RON.
For  VGG-19 pruned with DCP~\protect\cite{zhuang2018discrimination} approach at $0.3\%$ pruning rate, RON provides $4.48\times$ FLOP reduction with $0.27\%$ accuracy increase comparing to the initial VGG-19 while maintaining higher accuracy and better acceleration than the pruned baseline~(Figure~\ref{fig:cifar10}).

\begin{table}[]
    \centering
    \begin{tabular}{c|ccccccc}
\thead{Model \\ ~ \\ ~} & \thead{Modified \\ layers \\ ~} & \thead{Acc@1 \\ without \\ fine-tuning} & \thead{Acc@5 \\ without \\ fine-tuning} &  \thead{Acc@1 \\ with \\ fine-tuning} &  \thead{Acc@5 \\ with \\ fine-tuning} &  \thead{Speed up \\ on CPU \\ ~} & \thead{FLOP \\ reduction \\ ~} \\
\hline
  Teacher & --- &  --- &  --- &  71.95 &  89.41 &  1.00$\times$ &  1.00$\times$ \\
  RON $10 \times$ & 8 to 16 & 70.81 &  88.51 &  \textbf{72.09} &  \textbf{90.12} &  \textbf{1.95}$\times$ & \textbf{1.66} $\times$ \\
  RON $20 \times$ &  8 to 16 & 63.94 &  85.12 &  71.89 &  89.95 &  2.15$\times$ & 1.71$\times$\\
  RON $10 \times$ & 10 to 16 &  60.68 &  82.36 &  70.87 &  90.46 &  1.72$\times$ & 1.84$\times$\\
  RON $20 \times$ & 10 to 16 &  44.07 &  68.29 &  69.69 &  89.78 & 2.19$\times$ & 2.19$\times$ \\
 RON $10 \times$ & 12 to 16 & 42.77 &  67.34 &  66.84 &  88.16 &  2.22$\times$ & 2.58$\times$
 \end{tabular}
    \caption{VGG on CIFAR-100. RON $N\times$ stands for the accelerated model, where feature dimentionality of last layers is reduced by $N\times$ times comparing to the teacher.}
    \label{tab:cifar100}
\end{table}

\begin{table}[]
    \centering
    \begin{tabular}{c|ccccccc}
\textbf{Model} & \thead{Modified \\ layers} & \thead{Acc@1 without \\ fine-tuning} &  \thead{Acc@1 with \\ fine-tuning} &  \thead{Speed up \\ on CPU} & \thead{FLOP \\ reduction} \\
\hline
Teacher & --- & --- & 96.03 & 1.00$\times$ &  1.00$\times$ \\
RON $10 \times$ & 5 to 7 & 92.46 & 95.41 & 1.62$\times$ & 1.30$\times$ \\
RON $20 \times$ & 5 to 7 & 89.04 & 95.33 & 1.71$\times$ & 1.53$\times$ \\
RON $20 \times$ & 3 to 7 & 83.58 & 92.13 & 1.67$\times$ & 1.65$\times$ \\
\end{tabular}
    \caption{VGG on SVHN. RON $N\times$ stands for the accelerated model, where feature dimentionality of last layers is reduced by $N\times$ times comparing to the teacher.}
    \label{tab:svhn}
\end{table}




\subsection{Comparisons with other approaches}
\label{sec::compare}

The advantage of our method is that it can be applied both alone and on top of pruning algorithms. 
We have aggregated our results (Table~\ref{tab:my_label}) with the information from the paper by Zhuang et al.~\protect\cite{zhuang2018discrimination} and present them in Table~\ref{table:comparision_cifar10}. 
We compare RON with ThiNet~\protect\cite{ThiNet}, channel pruning (CP)~\protect\cite{he2017channelpruning}, network slimming~\protect\cite{Liu2017learning} and width-multiplier method~\protect\cite{wm}. 
More details about related methods can be found in Section~\ref{related_work}.

\begin{table}[ht!]
\centering
\begin{tabular}{l|cccc}
     \thead{Model \\ ~} &  \thead{FLOP \\ reduction} & \thead{Accuracy \\ drop, \%} \\
     \hline
     ThiNet \protect\cite{ThiNet} & 2.00$\times$ & 0.14\\
     Network Sliming  \protect\cite{Liu2017learning} & 2.04$\times$ & 0.19 \\
     Channel Pruning  \protect\cite{he2017channelpruning} & 2.00$\times$ & 0.32\\ 
     Width-multiplier  \protect\cite{wm} & 2.00$\times$ & 0.38\\
     Discrimination-aware Channel Pruning (DCP) \protect\cite{zhuang2018discrimination} & 2.00$\times$ & -0.17 \\
     DCP-Adapt  \protect\cite{zhuang2018discrimination} & 2.86$\times$ & -0.58 \\
     \hline
     RON (modified layers: 7 to 16) + fine-tuning  & {\bf 2.30}$\times$ & {\bf-0.18} \\ 
     DCP + RON (modified layers: 9 to 16) + fine-tuning & {\bf 3.37}$\times$ & {\bf-0.57} \\
     DCP + RON (modified layers: 7 to 16) + fine-tuning & {\bf 4.48}$\times$ & {\bf -0.27}\\
\end{tabular}
\caption{Comparison of acceleration methods for VGG-19 on CIFAR-10. Pre-trained baseline has $93.7\%$ accuracy. The higher FLOP reduction the better. The smaller accuracy drop the better.}
\label{table:comparision_cifar10}
\end{table}
\section{Discussion}
We have proposed a method that exploits the low-rank property of the outputs of neural network layers. 
The advantage of our approach is the ability to work with a large class of modern neural networks and obtain a simple fully-connected student neural network. 
We showed that, in some cases, the student model has the same quality as a student network even without any fine-tuning. 

The disadvantage of the Reduced-Order Network is that the number of parameters may increase when applied to wide convolutional networks on high-resolution images since the resulting network is dense. 
However, we have demonstrated that our method works well for neural networks with pruned channels, and such pruning allows us to reduce the number of features. 
The best application of our approach, in our opinion, is to further accelerate networks, which were produced by channel pruning algorithms. 

Later on, we can try sparsification~\cite{molchanov2017variational} and quantization techniques on top of our approach to mitigate this issue. 

\section{Conclusion}
We have developed a neural network inference acceleration method that is based on mapping layer outputs to a low-dimensional subspace using the singular value decomposition and the rectangular maximum volume algorithm. 
We demonstrated empirically that our approach allows finding a good initial approximation in the space of new model parameters. 
Namely, on CIFAR-10 and CIFAR-100, we achieved accuracy on par or even slightly better than the teacher model without fine-tuning and reached acceleration up to $4.48\times$ with fine-tuning and no accuracy drop. 
We have supported our experiments with the theoretical results, including approximation error upper bound evaluation.

\section{Acknowledgements}
This study was supported by RFBR, project number 19-31-90172 and 20-31-90127 (algorithm) and by the Ministry of Education and Science of the Russian Federation (grant 14.756.31.0001) (experiments).

\bibliographystyle{unsrt}
\bibliography{papers}

\begin{thebibliography}{10}

\bibitem{chen2018neural}
Tian~Qi Chen, Yulia Rubanova, Jesse Bettencourt, and David~K Duvenaud.
\newblock {Neural Ordinary Differential Equations}.
\newblock In {\em Advances in Neural Information Processing Systems}, pages
  6572--6583, 2018.

\bibitem{grathwohl2018ffjord}
Will Grathwohl, Ricky~TQ Chen, Jesse Betterncourt, Ilya Sutskever, and David
  Duvenaud.
\newblock {FFJORD: Free-form Continuous Dynamics for Scalable Reversible
  Generative Models}.
\newblock {\em arXiv preprint arXiv:1810.01367}, 2018.

\bibitem{gusak2020towards}
Julia Gusak, Larisa Markeeva, Talgat Daulbaev, Alexandr Katrutsa, Andrzej
  Cichocki, and Ivan Oseledets.
\newblock {Towards Understanding Normalization in Neural ODEs}.
\newblock {\em International Conference on Learning Representations (ICLR)
  Workshop on Integration of Deep Neural Models and Differential Equations},
  2020.

\bibitem{daulbaev2020interpolated}
Talgat Daulbaev, Alexandr Katrutsa, Larisa Markeeva, Julia Gusak, Andrzej
  Cichocki, and Ivan Oseledets.
\newblock Interpolation technique to speed up gradients propagation in neural
  odes.
\newblock {\em Advances in Neural Information Processing Systems}, 33, 2020.

\bibitem{quarteroni2014reduced}
Alfio Quarteroni, Gianluigi Rozza, et~al.
\newblock {\em Reduced order methods for modeling and computational reduction},
  volume~9.
\newblock Springer, 2014.

\bibitem{chaturantabut2010nonlinear}
Saifon Chaturantabut and Danny~C Sorensen.
\newblock Nonlinear model reduction via discrete empirical interpolation.
\newblock {\em SIAM Journal on Scientific Computing}, 32(5):2737--2764, 2010.

\bibitem{Cheng2018}
Yu~Cheng, Duo Wang, Pan Zhou, and Tao Zhang.
\newblock {Model Compression and Acceleration for Deep Neural Networks: The
  Principles, Progress, and Challenges}.
\newblock {\em IEEE Signal Processing Magazine}, 35(1):126--136, 2018.

\bibitem{Bucilua2006}
Cristian Bucilu\v{a}, Rich Caruana, and Alexandru Niculescu-Mizil.
\newblock Model compression.
\newblock In {\em Proceedings of the 12th ACM SIGKDD International Conference
  on Knowledge Discovery and Data Mining}, KDD '06, pages 535--541, New York,
  NY, USA, 2006. ACM.

\bibitem{Hinton2015}
Geoffrey Hinton, Oriol Vinyals, and Jeff Dean.
\newblock {Distilling the Knowledge in a Neural Network}.
\newblock {\em arXiv preprint arXiv:1503.02531}, 2015.

\bibitem{Romero2014}
Adriana Romero, Nicolas Ballas, Samira~Ebrahimi Kahou, Antoine Chassang, Carlo
  Gatta, and Yoshua Bengio.
\newblock Fitnets: Hints for thin deep nets.
\newblock {\em arXiv preprint arXiv:1412.6550}, 2014.

\bibitem{Zagoruyko2016_attention}
Sergey Zagoruyko and Nikos Komodakis.
\newblock {Paying More Attention to Attention: Improving the Performance of
  Convolutional Neural Networks via Attention Transfer}.
\newblock {\em arXiv preprint arXiv:1612.03928}, dec 2016.

\bibitem{li2016pruning}
Hao Li, Asim Kadav, Igor Durdanovic, Hanan Samet, and Hans~Peter Graf.
\newblock Pruning filters for efficient convnets.
\newblock {\em arXiv preprint arXiv:1608.08710}, 2016.

\bibitem{hu2016network}
Hengyuan Hu, Rui Peng, Yu-Wing Tai, and Chi-Keung Tang.
\newblock Network trimming: A data-driven neuron pruning approach towards
  efficient deep architectures.
\newblock {\em arXiv preprint arXiv:1607.03250}, 2016.

\bibitem{Liu2017learning}
Zhuang Liu, Jianguo Li, Zhiqiang Shen, Gao Huang, Shoumeng Yan, and Changshui
  Zhang.
\newblock Learning efficient convolutional networks through network slimming.
\newblock In {\em ICCV}, 2017.

\bibitem{wen2016learning}
Wei Wen, Chunpeng Wu, Yandan Wang, Yiran Chen, and Hai Li.
\newblock Learning structured sparsity in deep neural networks.
\newblock In {\em Advances in neural information processing systems}, pages
  2074--2082, 2016.

\bibitem{he2017channelpruning}
Yihui He, Xiangyu Zhang, and Jian Sun.
\newblock Channel pruning for accelerating very deep neural networks.
\newblock In {\em The IEEE International Conference on Computer Vision (ICCV)},
  Oct 2017.

\bibitem{ThiNet}
J.~{Luo}, H.~{Zhang}, H.~{Zhou}, C.~{Xie}, J.~{Wu}, and W.~{Lin}.
\newblock Thinet: Pruning cnn filters for a thinner net.
\newblock {\em IEEE Transactions on Pattern Analysis and Machine Intelligence},
  pages 1--1, 2018.

\bibitem{zhong2018prune}
Jing Zhong, Guiguang Ding, Yuchen Guo, Jungong Han, and Bin Wang.
\newblock Where to prune: Using lstm to guide end-to-end pruning.
\newblock In {\em IJCAI}, pages 3205--3211, 2018.

\bibitem{zhuang2018discrimination}
Zhuangwei Zhuang, Mingkui Tan, Bohan Zhuang, Jing Liu, Yong Guo, Qingyao Wu,
  Junzhou Huang, and Jinhui Zhu.
\newblock Discrimination-aware channel pruning for deep neural networks.
\newblock In S.~Bengio, H.~Wallach, H.~Larochelle, K.~Grauman, N.~Cesa-Bianchi,
  and R.~Garnett, editors, {\em Advances in Neural Information Processing
  Systems 31}, pages 881--892. Curran Associates, Inc., 2018.

\bibitem{Denton2014}
Emily~L Denton, Wojciech Zaremba, Joan Bruna, Yann LeCun, and Rob Fergus.
\newblock Exploiting linear structure within convolutional networks for
  efficient evaluation.
\newblock In {\em Advances in neural information processing systems}, pages
  1269--1277, 2014.

\bibitem{Jaderberg2014}
Max Jaderberg, Andrea Vedaldi, and Andrew Zisserman.
\newblock {Speeding up Convolutional Neural Networks with Low Rank Expansions}.
\newblock {\em arXiv preprint arXiv:1405.3866}, may 2014.

\bibitem{Lebedev2014}
Vadim Lebedev, Yaroslav Ganin, Maksim Rakhuba, Ivan Oseledets, and Victor
  Lempitsky.
\newblock {Speeding-up Convolutional Neural Networks Using Fine-tuned
  CP-Decomposition}.
\newblock {\em arXiv preprint arXiv:1412.6553}, dec 2014.

\bibitem{Zhang2015}
Xiangyu Zhang, Jianhua Zou, Kaiming He, and Jian Sun.
\newblock Accelerating very deep convolutional networks for classification and
  detection.
\newblock {\em IEEE transactions on pattern analysis and machine intelligence},
  38(10):1943--1955, 2015.

\bibitem{gusak2019automated}
Julia Gusak, Maksym Kholiavchenko, Evgeny Ponomarev, Larisa Markeeva, Philip
  Blagoveschensky, Andrzej Cichocki, and Ivan Oseledets.
\newblock Automated multi-stage compression of neural networks.
\newblock In {\em Proceedings of the IEEE International Conference on Computer
  Vision Workshops}, pages 0--0, 2019.

\bibitem{cui2019active}
Chunfeng Cui, Kaiqi Zhang, Talgat Daulbaev, Julia Gusak, Ivan Oseledets, and
  Zheng Zhang.
\newblock Active subspace of neural networks: Structural analysis and universal
  attacks.
\newblock {\em arXiv preprint arXiv:1910.13025}, 2019.

\bibitem{Courbariaux2014}
Matthieu Courbariaux, Yoshua Bengio, and Jean-Pierre David.
\newblock {Training deep neural networks with low precision multiplications}.
\newblock {\em arXiv preprint arXiv:1412.7024}, dec 2014.

\bibitem{Gupta2015}
Suyog Gupta, Ankur Agrawal, Kailash Gopalakrishnan, and Pritish Narayanan.
\newblock {Deep Learning with Limited Numerical Precision}.
\newblock In {\em International Conference on Machine Learning}, pages
  1737--1746, 2015.

\bibitem{mikhalev2018rectangular}
Aleksandr Mikhalev and Ivan~V Oseledets.
\newblock Rectangular maximum-volume submatrices and their applications.
\newblock {\em Linear Algebra and its Applications}, 538:187--211, 2018.

\bibitem{fonarev2016efficient}
Alexander Fonarev, Alexander Mikhalev, Pavel Serdyukov, Gleb Gusev, and Ivan
  Oseledets.
\newblock Efficient rectangular maximal-volume algorithm for rating elicitation
  in collaborative filtering.
\newblock In {\em 2016 IEEE 16th International Conference on Data Mining
  (ICDM)}, pages 141--150. IEEE, 2016.

\bibitem{woodruff2014sketching}
David~P Woodruff et~al.
\newblock Sketching as a tool for numerical linear algebra.
\newblock {\em Foundations and Trends{\textregistered} in Theoretical Computer
  Science}, 10(1--2):1--157, 2014.

\bibitem{tsitsulin2020frede}
Anton Tsitsulin, Marina Munkhoeva, Davide Mottin, Panagiotis Karras, Ivan
  Oseledets, and Emmanuel M{\"u}ller.
\newblock Frede: Linear-space anytime graph embeddings.
\newblock {\em arXiv preprint arXiv:2006.04746}, 2020.

\bibitem{he2016deep}
Kaiming He, Xiangyu Zhang, Shaoqing Ren, and Jian Sun.
\newblock Deep residual learning for image recognition.
\newblock In {\em Proceedings of the IEEE conference on computer vision and
  pattern recognition}, pages 770--778, 2016.

\bibitem{zagoruyko2016wide}
Sergey Zagoruyko and Nikos Komodakis.
\newblock Wide residual networks.
\newblock {\em arXiv preprint arXiv:1605.07146}, 2016.

\bibitem{huang2017densely}
Gao Huang, Zhuang Liu, Laurens Van Der~Maaten, and Kilian~Q Weinberger.
\newblock Densely connected convolutional networks.
\newblock In {\em Proceedings of the IEEE conference on computer vision and
  pattern recognition}, pages 4700--4708, 2017.

\bibitem{nakajima2013global}
Shinichi Nakajima, Masashi Sugiyama, S~Derin Babacan, and Ryota Tomioka.
\newblock Global analytic solution of fully-observed variational bayesian
  matrix factorization.
\newblock {\em Journal of Machine Learning Research}, 14(Jan):1--37, 2013.

\bibitem{simonyan2014very}
Karen Simonyan and Andrew Zisserman.
\newblock Very deep convolutional networks for large-scale image recognition.
\newblock {\em arXiv preprint arXiv:1409.1556}, 2014.

\bibitem{gao2018dynamic}
Xitong Gao, Yiren Zhao, Łukasz Dudziak, Robert Mullins, and Cheng zhong Xu.
\newblock Dynamic channel pruning: Feature boosting and suppression.
\newblock In {\em International Conference on Learning Representations}, 2019.

\bibitem{wm}
Andrew G.~Howard, Menglong Zhu, Bo~Chen, Dmitry Kalenichenko, Weijun Wang,
  Tobias Weyand, Marco Andreetto, and Hartwig Adam.
\newblock Mobilenets: Efficient convolutional neural networks for mobile vision
  applications.
\newblock {\em arXiv preprint arXiv:1704.04861}, 04 2017.

\bibitem{molchanov2017variational}
Dmitry Molchanov, Arsenii Ashukha, and Dmitry Vetrov.
\newblock Variational dropout sparsifies deep neural networks.
\newblock In {\em Proceedings of the 34th International Conference on Machine
  Learning-Volume 70}, pages 2498--2507. JMLR. org, 2017.

\end{thebibliography}
\end{document}